%% file: recomind.tex
\documentclass[sigconf,nonacm]{acmart}

\usepackage{algorithmic}
\usepackage{algorithm}  
\usepackage{graphicx}
\usepackage{textcomp}
\usepackage{xcolor}
\usepackage{bbm}
\input{macro.tex}
\AtBeginDocument{%
  }

\setcopyright{acmlicensed}
\copyrightyear{2018}
\acmYear{2018}
\acmDOI{XXXXXXX.XXXXXXX}

\acmISBN{978-1-4503-XXXX-X/2018/06}

\begin{document}

\title{RecoMind: A Reinforcement Learning Framework for Optimizing In-Session User Satisfaction in Recommendation Systems}  

\author{Mehdi Ben Ayed}
\email{mbenayed@pinterest.com}
\orcid{0009-0000-5727-3219}
\affiliation{%
 \institution{Pinterest Inc.}
  \city{New York}
  \country{USA}
}

\author{Fei Feng}
\email{ffeng@pinterest.com}
\affiliation{%
  \institution{Pinterest Inc.}
  \city{San Francisco}
  \country{USA}
}

\author{Jay Adams}
\email{jadams@pinterest.com}
\affiliation{%
  \institution{Pinterest Inc.}
  \city{San Francisco}
  \country{USA}
}

\author{Vishwakarma Singh}
\email{vishwakarmasingh@pinterest.com}
\affiliation{%
  \institution{Pinterest Inc.}
  \city{San Francisco}
  \country{USA}
}

\author{Kritarth Anand}
\email{kanand@pinterest.com}
\affiliation{%
 \institution{Pinterest Inc.}
  \city{San Francisco}
  \country{USA}
}

\author{Jiajing Xu}
\email{jiajing@pinterest.com}
\affiliation{%
  \institution{Pinterest Inc.}
  \city{San Francisco}
  \country{USA}
}

\renewcommand{\shortauthors}{Ben Ayed et al.}
\input{abstract}



\maketitle

\input{intro}

\input{relatedwork}

\input{mdp.tex}

\input{framework.tex}

\input{experiment.tex}

\input{conclusion.tex}

\bibliographystyle{ACM-Reference-Format}
\bibliography{recomind}

\end{document}

%% file: macro.tex
\usepackage{mathtools}
\usepackage{graphicx}
\usepackage{amsthm}
\usepackage{xpatch}
\usepackage{xspace}
\usepackage{mathrsfs}

\usepackage{url}
\usepackage{array}
\usepackage{wrapfig}
\usepackage{multirow}
\usepackage{tabularx}
\usepackage[normalem]{ulem} 
\usepackage{enumerate}
\usepackage{enumitem}

\newcommand\cut[1]{{}}  
\makeatletter
\xpatchcmd{\algorithmic}{\itemsep\z@}{\itemsep=.5ex plus.5pt}{}{}
\makeatother
\usepackage{mathtools}
\mathtoolsset{showonlyrefs}

\newcommand{\vs}{{\mathbf{s}}}

\newcommand{\cA}{{\mathcal{A}}}

\newcommand{\cF}{{\mathcal{F}}}

\newcommand{\cM}{{\mathcal{M}}}

\newcommand{\cS}{{\mathcal{S}}}

\newcommand{\cU}{{\mathcal{U}}}

\newcommand{\RR}{\mathbb{R}}
\newcommand{\PP}{\mathbb{P}}

\renewcommand{\iota}{I}

\DeclareMathOperator*{\argmax}{argmax}

\newcommand{\bc}{\begin{center}}
\newcommand{\ec}{\end{center}}

\newcommand{\bdm}{\begin{displaymath}}
\newcommand{\edm}{\end{displaymath}}

\newcommand{\beq}{\begin{equation}}
\newcommand{\eeq}{\end{equation}}

\newcommand{\bfl}{\begin{flushleft}}
\newcommand{\efl}{\end{flushleft}}

\newcommand{\bt}{\begin{tabbing}}
\newcommand{\et}{\end{tabbing}}

\newcommand{\beqn}{\begin{align}}
\newcommand{\eeqn}{\end{align}}

\newcommand{\beqs}{\begin{align*}} 
\newcommand{\eeqs}{\end{align*}}  

\providecommand{\keywords}[1]
{
  \small	
  \textbf{\textit{Keywords---}} #1
}

\numberwithin{theorem}{section}

\numberwithin{assumption}{section}

\newtheorem{remark}{Remark}

\numberwithin{lemma}{section}

\makeatletter
\renewcommand*\env@matrix[1][\arraystretch]{%
  \edef\arraystretch{#1}%
  \hskip -\arraycolsep
  \let\@ifnextchar\new@ifnextchar
  \array{*\c@MaxMatrixCols c}}
\makeatother

\newenvironment{itemize*}%
{\begin{itemize}[leftmargin=*,topsep=5pt]%
		\setlength{\itemsep}{1pt}%
		\setlength{\parskip}{1pt}}%
	{\end{itemize}}

\newenvironment{enumerate*}%
{\begin{enumerate}[leftmargin=20pt,topsep=5pt]%
		\setlength{\itemsep}{1pt}%
		\setlength{\parskip}{1pt}}%
	{\end{enumerate}}

\makeatletter
\DeclareRobustCommand\onedot{\futurelet\@let@token\@onedot}
\def\@onedot{\ifx\@let@token.\else.\null\fi\xspace}

 \def\vs{\emph{vs}\onedot}

\makeatother

%% file: abstract.tex
\begin{abstract}
Existing web-scale recommendation systems commonly use supervised learning methods that prioritize immediate user feedback. 
Although reinforcement learning (RL) offers a solution to optimize longer-term goals, such as in-session engagement, applying it at web scale is challenging due to the extremely large action space and engineering complexity. In this paper, we introduce RecoMind, a simulator-based RL framework designed for the effective optimization of session-based goals at web-scale. RecoMind leverages existing recommendation models to establish a simulation environment and to bootstrap the RL policy to optimize immediate user interactions from the outset. This method integrates well with existing industry pipelines, simplifying the training and deployment of RL policies. Additionally, RecoMind introduces a custom exploration strategy to efficiently explore web-scale action spaces with hundreds of millions of items. We evaluated RecoMind through extensive offline simulations and online A/B testing on a video streaming platform. Both methods showed that the RL policy trained using RecoMind significantly outperforms traditional supervised learning recommendation approaches in in-session user satisfaction. In online A/B tests, the RL policy increased videos watched for more than 10 seconds by 15.81\% and improved session depth by 4.71\% for sessions with at least 10 interactions. As a result, RecoMind presents a systematic and scalable approach for embedding RL into web-scale recommendation systems, showing great promise for optimizing session-based user satisfaction.
\end{abstract}

%% file: intro.tex
\vspace{-5pt} 
\section{Introduction}
Recommendation systems are omnipresent in a variety of domains, including movies, music, and social media \cite{das2017survey, goyani2020review, song2012survey, feng2020news, anandhan2018social}. These systems enhance the user experience by surfacing content based on user interests, thereby boosting user engagement and satisfaction. Traditional recommendation approaches, which are typically greedy one-step models, focus on maximizing immediate user satisfaction by taking into account the user context and recommending items with the highest prediction scores for actions such as clicks \cite{DLRM19, TransAct, DLRecSys22, GNNRecSys23, DLRecSys19, SSLRecSys23}. This method produces effective short-term gains, but can result in repetitive and narrow content due to a disregard for long-term user engagement factors, such as user stickiness and session metrics \cite{AndersonWWW20, clarke2008novelty}, where a session is defined as a series of temporally contiguous user interactions with the system.

In recent years, there has been a growing interest in long-term recommendation strategies driven largely by advances in reinforcement learning (RL). These strategies aim to extend the optimization horizon beyond immediate user actions, focusing on improving in-session and cross-session user satisfaction \cite{RLRecSys23, RLRecSys22, DLRSur21}. Unlike traditional greedy one-step models, such as neural network (NN)-based methods that predict immediate user feedback, RL policies are designed to optimize expected cumulative rewards, making them particularly well-suited to capture sustained user engagement over extended periods.
A body of empirical work has demonstrated the effectiveness of RL through offline tests \cite{JobSkillRL21, MusicRL20, CourseRL219, TreatRecRL18} and online experiments \cite{tomasi2023automatic,deffayet2023generative, topkoff19}. However, the deployment of RL to online web-scale recommendation systems remains scarce due to two major challenges \cite{chen2023deep, RLRecSys23, DLRSur21} that are yet to be addressed:
\begin{itemize}[leftmargin=15pt]  
\item \textbf{Exploration at web-scale}: Efficiently exploring hundreds of millions of items while ensuring computational and sample efficiency, as well as rapid convergence, is challenging.
\item \textbf{Significant engineering efforts to productionize RL}: Transitioning existing industrial pipelines from supervised to reinforcement learning is costly, risky, and resource-intensive.
\end{itemize}  

In this paper, we propose RecoMind, a RL framework for web-scale recommendation systems. RecoMind addresses key challenges in deploying and running RL at web-scale by providing an approach to transition existing supervised learning (SL) pipelines to reinforcement learning and by introducing mechanisms to enable exploration at this vast scale.

One of the primary strengths of RecoMind is its compatibility with industrial pipelines that rely on supervised learning (SL). Traditional RL methods often require modifications to production infrastructure due to specialized model architectures and the potential need for custom simulators for training \cite{RLRecSys22}. In contrast, RecoMind seamlessly adopts the same interfaces as existing SL models at both the feature and output levels. This allows it to effectively use the current ML infrastructure without any alterations. Furthermore, RecoMind enables the use of existing SL models to power the simulation component directly, eliminating the necessity of building custom simulations from scratch. This alignment minimizes disruptions and reduces the engineering effort required to adapt existing systems for RL deployment, enabling the development of competitive RL methods within weeks, rather than months.

Another critical feature of RecoMind is its ability to bootstrap the value network (or policy) of the RL agent using an existing greedy one-step model, embedding an immediate understanding of short-term action impacts. The motivations behind transfer learning from a production model are two-fold. First, initializing with the knowledge of the production model accelerates the RL training process by providing a more effective initial policy. Second, the production model's established track record of reliable performance enhances the safety of deploying an RL policy by ensuring that it remains close to the proven production behavior.

Additionally, RecoMind introduces an innovative exploration strategy optimized to effectively navigate web-scale action spaces. Traditional methods like epsilon-greedy and SoftQ struggle with large action spaces: epsilon-greedy fails due to extreme reward sparsity from many low-reward actions, and SoftQ is computationally expensive and hard to tune. RecoMind combines these methods by primarily using epsilon-greedy exploration while applying SoftQ to a truncated set of top-$K$ $Q$-value actions during the exploration phase. This custom strategy focuses on high-value actions while maintaining diversity, thereby improving convergence speed and exploration efficiency in web-scale recommendation systems. Notably, RecoMind successfully operates in an action space containing 100 million items, a scale at which conventional methods typically struggles due to computational and tuning challenges.

We conducted extensive offline and online experiments comparing RecoMind with the widely used supervised learning approach to assess its effectiveness in improving recommendation performance.  Offline tests show a performance improvement of the RL policy over the greedy one-step model, which selects items with highest immediate predicted score. This improvement is supported by a series of ablation studies demonstrating the impacts of each technique proposed in this paper. In the online experiment, we deploy the RL policy on a video streaming platform, observing significant engagement gains that align with the offline results. These results highlight RecoMind's potential and demonstrate RL's effectiveness in improving in-session metrics, a crucial step towards long-term optimization. In summary, our contributions are:
\begin{itemize}[topsep=0pt, leftmargin=15pt]
    \item \textbf{RecoMind Framework:} We present RecoMind, an innovative RL framework designed for web-scale recommendation systems to achieve in-session optimization.
    \item \textbf{Facilitating Transition to RL:} 
    RecoMind facilitates the transition from supervised to reinforcement learning by using existing the supervised learning model for simulation and bootstrapping the RL policy, while using existing infrastructure for seamless deployment.

    \item \textbf{Web-Scale Exploration Technique:} RecoMind introduces an innovative exploration strategy specifically designed for web-scale action spaces, effectively addressing the challenges of exploration efficiency and reward sparsity.

    \item \textbf{Benefits in Offline/Online Experiments:} 
    We validate RecoMind through offline and online experiments, showing improvements in key metrics on a video streaming platform.   
\end{itemize}  
\vspace{-5pt} 

%% file: relatedwork.tex
\vspace{-5pt} 
\section{Related Work}
Traditional recommender systems have focused primarily on optimizing instant metrics, such as clicks, to improve immediate user satisfaction, often relying on greedy one-step models. \cite{chang2017streaming, hidasi2015session, li2017neural, mnih2007probabilistic}. These approaches encompass a variety of techniques, including collaborative filtering and deep neural networks like MLPs, CNNs, RNNs, and attention architectures \cite{chang2017streaming, cheng2016wide, wu2016collaborative, bai2019ctrec, gu2020hierarchical, bai2018attribute}. However, these metrics fall short in capturing long-term user engagement (e.g., user stickiness, session depth, etc.), leading to myopic policies that fail to account for iterative interactions, in-session dynamics, and delayed rewards. Studies have also highlighted the detrimental effects of this short-term focus, including decreased user retention, and the rich-get-richer issue, which ultimately harm both users and content providers \cite{clarke2008novelty, anderson2020algorithmic, chen2019top, hohnhold2015focusing, waller2019generalists}. Although there have been attempts to mitigate these issues by improving diversity or identifying metrics correlated with long-term results, these methods do not directly optimize session-level interactions, which are crucial for fostering long-term engagement \cite{chandar2022using, zou2019reinforcement}.

To address these issues, RL provides a natural alternative to methods focused on optimizing immediate metrics \cite{dulac2015deep, lu2016partially, zhao2019deep, zhao2017deep, zou2019reinforcement}. By casting the problem as a Markov Decision Process (MDP), RL directly aims for long-term goals through maximizing cumulative rewards. Bandit algorithms were first widely adopted to control long-term regrets, but while they handle the exploration-exploitation dilemma, they neglect inter-step correlation, which can lead to suboptimality for dynamic platforms \cite{he2019off, li2010contextual, wang2017factorization, zeng2016online}. Following this, there were attempts to leverage online RL for dynamic environments, but training RL directly in an online environment could result in a degradation of user experience [34]. Subsequently, offline batch RL emerged as an alternative, but it can lead to sub-optimal policies due to data coverage limitations \cite{qin2014contextual, wang2017factorization, xiangyu2017deep, swaminathan2015counterfactual, chen2019top}.

Research in robotics has effectively tackled sparse reward environments and transitioned from simulations to real-world applications \cite{andrychowicz2017hindsight}. Notably, \cite{nagabandi2018neural} introduced a hybrid model-based and model-free RL approach, which considerably improved sample efficiency in robotic locomotion. However, this integration was tested within the limited action spaces common to robotics. 

In the context of recommendation systems, researchers \cite{tomasi2023automatic, deffayet2023generative} have trained RL agents for recommendation systems using simulated environments. GoogleRecSim \cite{ie2019recsim} is another noteworthy tool in this domain; it provides a configurable environment specifically designed to evaluate and research reinforcement learning algorithms for recommendation systems. However, these efforts operated within much smaller action spaces. Moreover, these efforts were limited to smaller action spaces and did not address the extensive engineering challenges and crucial transition strategies needed for shifting from greedy one-step methods to RL methods.

In this paper, the proposed framework, RecoMind, distinguishes itself by leveraging existing supervised learning methods to facilitate the transition to RL and by introducing an innovative exploration strategy tailored for web-scale action spaces.

%% file: mdp.tex
\vspace{-5pt} 
\section{Problem Formulation}\label{sec:mdp}
Web-scale recommendation systems, such as feed view streaming platforms, aim to keep users engaged by providing personalized recommendations. To optimize for long-term user engagement, with a particular focus on in-session dynamics, we frame the recommendation problem as an RL task. Although RL requires exploration to learn effective strategies, performing it directly on real users can degrade their experience. Therefore, we develop a simulator to model direct user feedback, enabling offline exploration and learning.

We formulate the recommendation task as a Markov Decision Process (MDP), defined by $\cM = (\cS, \cA, P, R, \gamma)$.
$\cS$ represents the state space, $\cA$ is the action space, $P: \cS \times \cA \times \cS
\rightarrow \RR$ specifies the transition probability, $R: \cS \times \cA \rightarrow \RR$
is the reward function, and $\gamma \in \left[0, 1\right]$ is a discount factor. At each 
step, the agent observes a state $s$ and selects an action $a$ following a policy 
$\pi: \cS\rightarrow \cA$. The environment returns an immediate reward $R(s,a)$ and transitions to a new state $s'$ 
with probability $P(s,a,s')$. The agent's goal is
to learn a policy that maximizes the value function $V^{\pi}(s_0):=\mathbb{E}\big[\sum_{t=0}^{T} \gamma^t r_t\big]$.

In the context of recommender systems, interactions occur between a recommendation system and a user $u \in \cU$. At each time step $t$, the recommendation system presents the user with am item $a_t$ from a set of candidates based on user context. The user then provides explicit feedback $\cF_t$, which can include multiple responses such as save, hide, or exit. This feedback helps the recommendation system understand the user's preference. A sequence of these interactions constitute a "session", beginning when a user starts engaging and ending upon exit. Formally, we define the MDP as
\begin{itemize}[leftmargin=15pt]
    \item $\cS$: the state at time $t$ \begin{equation}
    s_t := \big\{u, (a_{t-L}, \cF_{t-L}), \dots, (a_{t-1}, \cF_{t-1})\big\},
    \end{equation}
     where $u$ is the user embedding that encapsulates a user's interests and $(a_{t-L}, \cF_{t-L})$, $\dots, (a_{t-1}, \cF_{t-1})$ contains the most recent $L$ recommended items with their user's feedback. 
    \item $\cA$: a set of hundreds of millions of recommendable items. 
    \item $R$:  a numerical reward based on diverse user feedback.
\end{itemize}
To capture the dynamic nature of user interactions, the transitions are grouped into distinct episodes (sessions). At the start of an episode, when $t=0$, we initialize $s_0$ with the user embedding and the most recent $L$ recommended items along with the user's feedback before the start of the session. To construct $s_{t+1}$, the feedback pair $(a_t, \mathcal{F}_t)$ is appended to recent interactions. If we already have $L$ feedback pairs, the oldest pair is removed. Each episode involves a fixed user and terminates when the user exits the platform.  

In summary, our goal is to optimize long-term user engagement  at the session level in web-scale recommendation systems by modeling the task as an MDP and using RL to develop policies that select actions with the highest expected cumulative value.

\begin{remark} In the definition of states, $L$ acts as a lookback window, influencing both policy performance and state space size.  A larger $L$ processes more user history but exponentially grows the state space. Practitioners should adjust this parameter as needed.
\end{remark}

%% file: framework.tex
\vspace{-5pt} 
\section{RecoMind Framework}\label{sec:framework}
In this section, we present the RecoMind framework. We start with a high-level overview, followed by the techniques used to apply RL to web-scale recommendation tasks. 
\vspace{-5pt} 
\subsection{High-level Structure}
\begin{figure}
    \vspace{-5pt} 
    \centering
\includegraphics[scale=0.12]{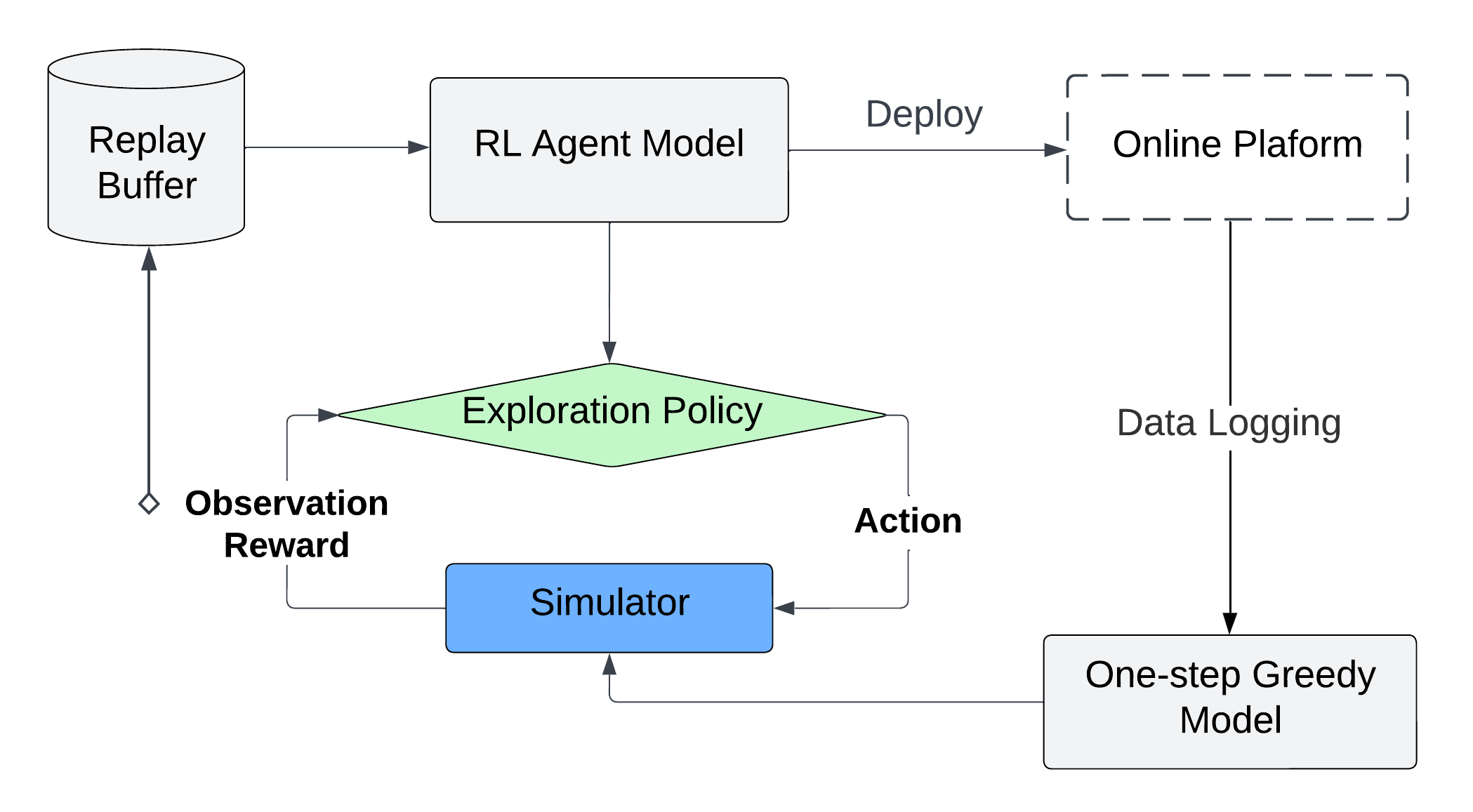}
    \vspace{-20pt} 
    \caption{The Integrated Loop Structure of the RecoMind Framework. This figure depicts the offline RL training cycle with a simulator, and the online deployment cycle using the RL policy for real-time user recommendations.}
    \label{fig:framework}
    \vspace{-15pt} 
\end{figure}

RecoMind has an integrated loop structure for offline training and online serving, ensuring a continuous improvement cycle for recommendation agents, as illustrated in Figure \ref{fig:framework}. This design leverages the existing one-step greedy model and fine-tunes it for RL training. 

The entire workflow can be broken down into two parts:

\begin{itemize}[leftmargin=15pt]
\item \textbf{Offline Training}
\vspace{3pt}
\begin{enumerate}[leftmargin=15pt]
\item Train a one-step greedy model using logged user interaction data to predict user feedback to recommended items.
\item Set up a simulator powered by the trained one-step greedy model to handle the transition function.
\item 
Train the RL agent by interacting with the simulator, optimizing the policy to maximize cumulative rewards.
\end{enumerate}
\vspace{5pt}
\item \textbf{Online Deployment}
\vspace{3pt}
\begin{enumerate}[leftmargin=15pt]
\item Deploy the trained policy to the online environment once the rewards stabilizes.
\item Log online user interactions and feedback during the serving phase to further enhance offline training.
\end{enumerate}
\end{itemize}
For more details on the offline training process, see Algorithm \ref{alg:sim_and_train}.  

We will now present the key contributions of RecoMind, highlighting how each component addresses the unique challenges of applying RL to web-scale recommendation systems.

\vspace{-5pt} 
\subsection{Leveraging Existing Recommender Systems}\label{sec:warm}
RecoMind leverages existing one-step greedy recommendation models, offering several key benefits. Firstly, it enables a smooth transition to the RL framework with minimal engineering overhead, reducing re-engineering efforts, costs, and risks. Additionally, pre-trained models expedite convergence and shorten training time. Finally, they provide a strong baseline performance, serving as a reliable benchmark during the transition from SL to RL. 

\begin{figure} 
    \vspace{-5pt} 
    \centering
    \includegraphics[scale=0.12]{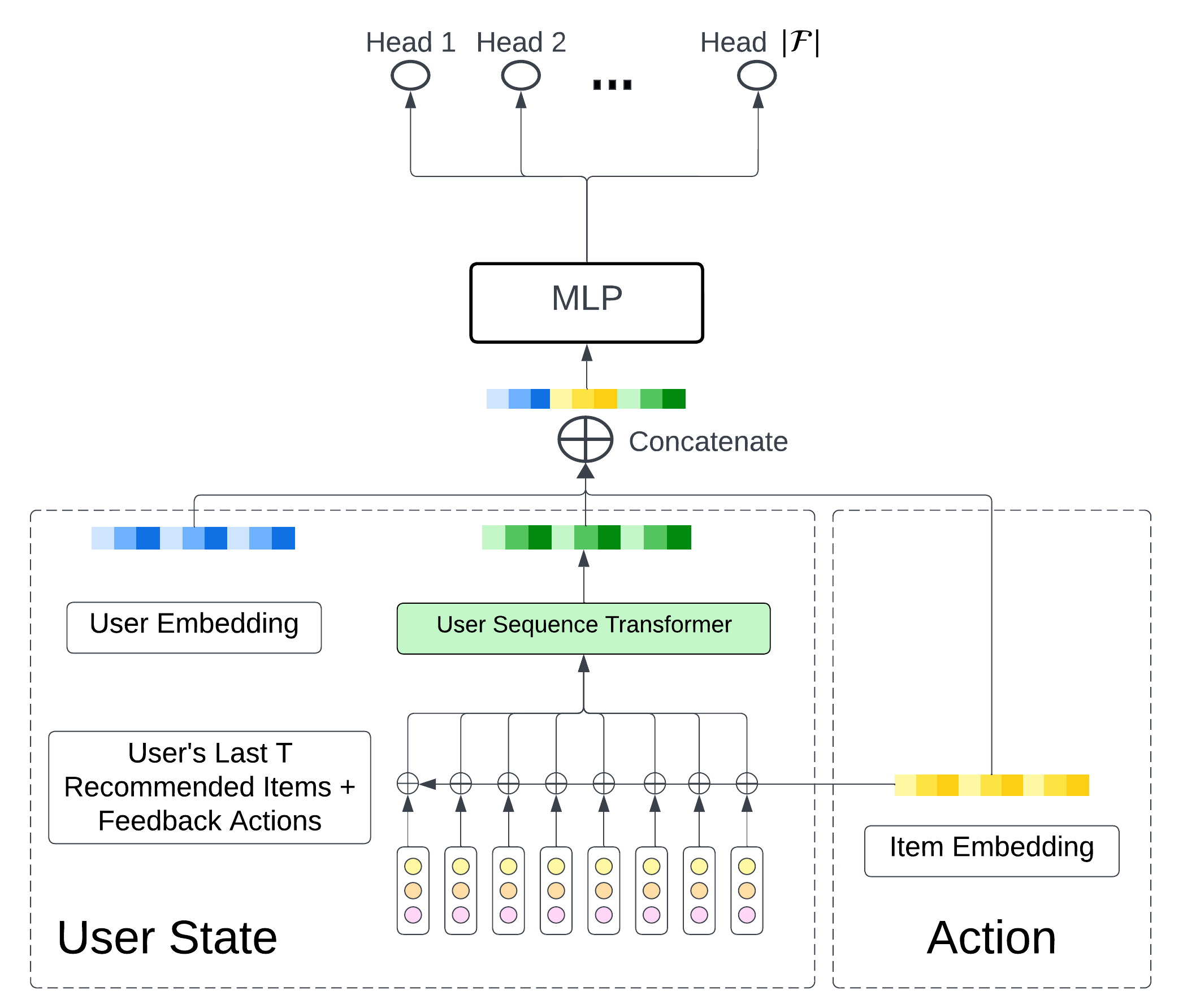}
    \vspace{-20pt} 
    \caption{The simulator architecture, adapted from an existing recommendation model. The simulator is a deep neural network that takes user state and action as input and outputs multi-head predictions.}
    \label{fig:simulator}
    \vspace{-15pt} 
\end{figure}

Central to our framework is a simulated environment, which uses predictions from the existing one-step greedy model to simulate user feedback. These predictions enable RecoMind to update the state $S$ with the most recent recommended items and their simulated feedback. Formally, if $s_t$ is the current state at time step $t$, $a_t$ is the action taken (i.e., the recommended item). The one-step model is then used to generate predicted feedback $\cF_t$ as shown in Figure \ref{fig:simulator} and the next state is updated as $s_{t+1}=s_t\cup\{a_t, \cF_t\}$ to reflects the latest simulated interactions. A weighted sum of all feedback predictions is also used to compute the reward $r_t$ for action $a_t$. The reward $r_t$ is defined as:   
\begin{align}\label{eq:reward}
r_t := \sum_{i=1}^{i=|\cF|} c_i \cdot \PP(f_i\in\cF_t ~|~ s_t, a_t),
\end{align}
where $c_i$ represents a weight associated with the feedback $f_i$, and $\PP(f_i\in\cF_t ~|~ s_t, a_t)$ is the probability of feedback $f_i$ given the state $s_t$ and action $a_t$, and $\cF$ is the full set of all possible feedback actions. 
\begin{remark} [Reward Design]
There are several merits of such a reward function:
\begin{itemize}[leftmargin=15pt]
\item Alignment with Existing Models: The rewards are output from the existing prediction model. To gain high rewards, the agent is encouraged to stay close to the preference of the existing model, ensuring that the RL policy does not deviate significantly from the current production behavior. This implicit conservatism is a positive property in offline RL, as it prevents drastic changes that could negatively impact user experience. 
\item Denser Reward Signals: The rewards are designed as prediction scores rather than 0/1 event indicators (e.g., the reward is 1 if the user watches the video and 0 if not). This denser signal overcomes the exploration difficulty allowing the RL model to learn more efficiently from fewer samples.
\item Comprehensive Evaluation: The reward function is a weighted sum that considers all possible feedback, allowing the model to balance multiple objectives. The weights in such a function can be set on the basis of the importance of the different actions relative to each other.
\end{itemize}
For more detailed insights and analysis on the impact of our reward design, please refer to the ablation study in Section \ref{sec:experiment}.
\end{remark}

A final advantage of using existing one-step greedy models lies in bootstrapping RL networks with the pre-trained supervised models. Specifically, we adopt the existing recommender's network architecture for RL models, where all layers are the same except for the output nodes. This pretraining boosts sample efficiency and ensures more stable convergence compared to learning from scratch. However, it is important to note a fundamental difference in the objectives of the $Q$-network compared to the existing model: While the existing model predicts immediate rewards, the $Q$-network predicts cumulative rewards over $T$ steps. Formally, this can be expressed as:
\begin{align}
Q(s,a) = \mathbb{E}\left[\sum_{t=0}^T \gamma^t r_t ~|~ (s_0, a_0)=(s,a)\right].
\end{align}

The $Q$-network is trained using Bellman optimality equation and the objective is to minimize the Temporal Difference (TD) error:
\begin{align}\label{eq:td_error}
L(\theta) = \mathbb{E}_{s, a, r, s'}\left[ \left( r + \gamma \max_{a'} Q(s', a'; \theta) - Q(s, a; \theta) \right)^2 \right],  
\end{align}  
where $\theta$ represents the network parameters.

In summary, leveraging existing one-step prediction models is central to RecoMind's design. This approach facilitates an efficient transition to RL with minimal engineering efforts, enables policy bootstrapping, and ensures reliable performance. In Section \ref{sec:experiment}, we compare the performance of using transfer-learning from the existing production model versus learning from scratch, demonstrating significant improvement with this approach. 

\vspace{-5pt} 
\subsection{RL Algorithm}\label{sec:rlalgorithm}
RecoMind enhances traditional RL algorithms to meet the demands of web-scale recommendation systems. It employs a state-action-in-value-out network to generalize the long-term value estimates of recommendations across a vast action space. Additionally, RecoMind introduces a novel exploration strategy that combines $\varepsilon$-greedy and softmax techniques, enabling more effective exploration. These features collectively enhance sample efficiency and scalability, enabling long-term optimization in web-scale environments.

\begin{figure}
    \vspace{-2pt} 
    \centering
    \includegraphics[scale=0.55]{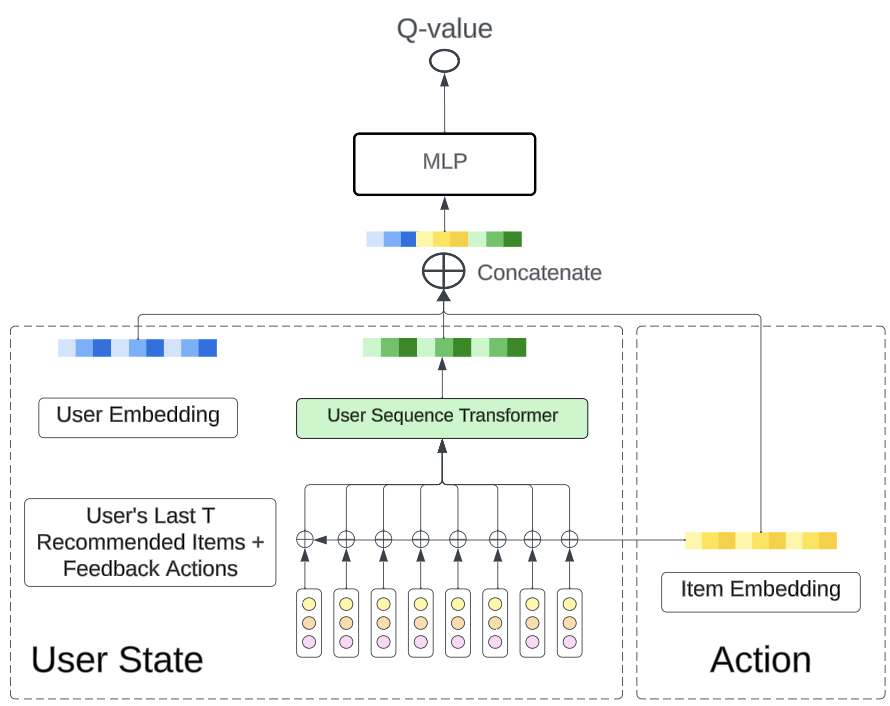}
    \caption{The $Q$-network architecture, adapted from an existing recommendation model. The $Q$-network is a deep neural network that includes an additional $|\cF|$-to-1 layer in the Multi-Layer Perceptron (MLP) to output a single $Q$-value.}
    \label{fig:DQN}
    \vspace{-5pt} 
\end{figure}

The state-action-in-value-out network is a key component in the RecoMind framework, enabling the model to effectively navigate the web-scale action space. In this network, the model takes both the state and action features as inputs and outputs the long-term value estimate of that action. This computation is performed for each action to evaluate all possible actions. This design overcomes the limitation of the traditional state-in-value-out $Q$-network, which fails to generalize across hundred of millions of items because of their lack of action features. See Figure \ref{fig:DQN} for a visual representation.

Given this value network, our objective is to optimize the policy defined as:    
$$    
\pi: \cS \rightarrow \cA,    
$$    
where the policy $\pi$ evaluates the state $s_t$ and considers the available actions $\cA$, selecting the action that yields the highest $Q$-value:    
$$    
\pi(s_t) = \arg\max_{a \in \cA} Q(s_t, a).    
$$    
To effectively train the $Q$-network, we minimize the Temporal Difference (TD) error, as discussed in Section \ref{sec:warm}. This training methodology ensures that the network learns to predict the long-term value of different recommendations.

To address the challenge of exploring with such a large action space, we develop an innovative exploration strategy, which overcomes the sample inefficiency of traditional methods such as vanilla $\varepsilon$-greedy and SoftQ exploration \cite{sutton2018reinforcement}. For example, $\varepsilon$-greedy exploration becomes inefficient in vast action spaces because it uniformly samples all possible actions, resulting in extremely sparse and often uninformative feedback. Similarly, pure softmax exploration, although more directed, can be computationally expensive and unstable, favoring suboptimal actions if the temperature parameter is not optimally tuned. This instability makes it particularly challenging to tune the temperature parameter during training. Specifically, our custom strategy is defined as follows:

\begin{equation}\label{eq:exploration}  
a_t \sim   
\begin{cases}   
\displaystyle \argmax_{a \in \cA} Q(s_t, a) & \!\!\!\!\!\!\!\!\!\text{with prob } 1 - \varepsilon, \\  
\displaystyle \frac{\exp(Q(s_t,a))}{\sum_{a'\in \cA^K_{Q}(s_t)} \exp(Q(s_t,a'))} \cdot \mathbbm{1}\left\{ a \in \cA^K_{Q}(s_t)\right\} & \text{with prob } \varepsilon,  
\end{cases}  
\end{equation}  

Using this strategy, the agent exploits with probability $1-\varepsilon$ by selecting the action with the highest $Q$-value. With probability $\varepsilon$, it explores by sampling from the top-$K$ actions (ranked by $Q$-value) using a softmax distribution. 

This approach ensures a balance between exploitation and targeted exploration. By truncating to the top-$K$ actions, the exploration focuses on the most promising candidates, mitigating the inefficiency of random exploration in a vast space. At the same time, softmax sampling within the truncated set ensures that exploration remains probabilistic and thus sufficiently diverse, while still preferring actions with higher estimated $Q$-values, leading to faster convergence.
The effectiveness of this custom exploration strategy compared to $\varepsilon$-greedy and softmax policies is empirically demonstrated in Section \ref{sec:exp_explore}. 

\begin{remark}
The values of $K$ and $\varepsilon$ determine the extent of exploration. Higher values increase the difficulty of exploration, while lower values can limit the quality of the policy. These parameters must be tuned to balance between exploration breadth and convergence speed.

\end{remark}

\vspace{-10pt} 
\subsection{Simulation and Training Algorithm}  
This section outlines recomind's training procedure which is detailed in Algorithm \ref{alg:sim_and_train}. It consistes of two main phases:

\begin{enumerate}  
    \item \textbf{Interaction with the Simulator:}  
    The agent interacts with the environment to generate new episodes, adding them to the replay buffer for training. This process uses a dataset of logged initial user states, obtained from the production environment, to initiate the simulated episodes.  
  
    \item \textbf{Policy Optimization:}  
    The policy optimization phase focuses on refining the $Q$-network based on the collected experience, minimizing the Temporal Difference (TD) error.  
\end{enumerate}  

For more details on the exploration policy and warm-starting, refer to Sections \ref{sec:framework} and \ref{sec:warm}. 

\begin{algorithm}
    \caption{Simulation and Training in RecoMind}      
    \label{alg:sim_and_train}      
    \begin{algorithmic}[1]      
        \STATE \textbf{Input:} Dataset $\mathcal{D}$, Greedy model parameters $\theta_{\text{greedy}}$  
        \STATE \textbf{Output:} Trained $Q$-network parameters $\theta_Q$  
        \STATE Initialize $Q$-network $\theta_Q$ with $\theta_{\text{greedy}}$, initialize replay buffer $\mathcal{B}$    
        \REPEAT      
            \FOR{$i = 1$ to $N_{\text{COLLECTION}}$}      
                \STATE Sample initial state $s_0 \sim \mathcal{D}$      
                \STATE $terminated \leftarrow \text{False}$
                \WHILE{not $terminated$}      
                    \STATE Select $a_t$ following exploration policy in Eq. \eqref{eq:exploration}     
                    \STATE Predict feedback $\cF_t$ given $(s_t, a_t)$ using greedy model
                    \STATE Update state $s_{t+1} \leftarrow s_t \cup \{a_t, \cF_t\}$      
                    \STATE Compute reward $r_t$ using Eq. \eqref{eq:reward}     
                    \STATE Store transition $\left(s_t, a_t, r_t, s_{t+1}\right)$ in $\mathcal{B}$ 
                    \STATE Update $terminated$ if session exit was predicted in $\cF_t$. 
                \ENDWHILE      
            \ENDFOR      
            \FOR{$j = 1$ to $N_{\text{TRAINING}}$}      
                \STATE Sample mini-batch from $\mathcal{B}$      
                \STATE Update $\theta_Q$ by minimizing TD-error (Eq. \eqref{eq:td_error})    
            \ENDFOR      
        \UNTIL{reward \text{stabilizes over an extended period}}    
    \end{algorithmic}      
\end{algorithm}   
\vspace{-5pt} 
\subsection{Distributed Architecture} \label{sec:distri}
\begin{figure}
    \vspace{-2pt} 
    \centering
\includegraphics[scale=0.43]{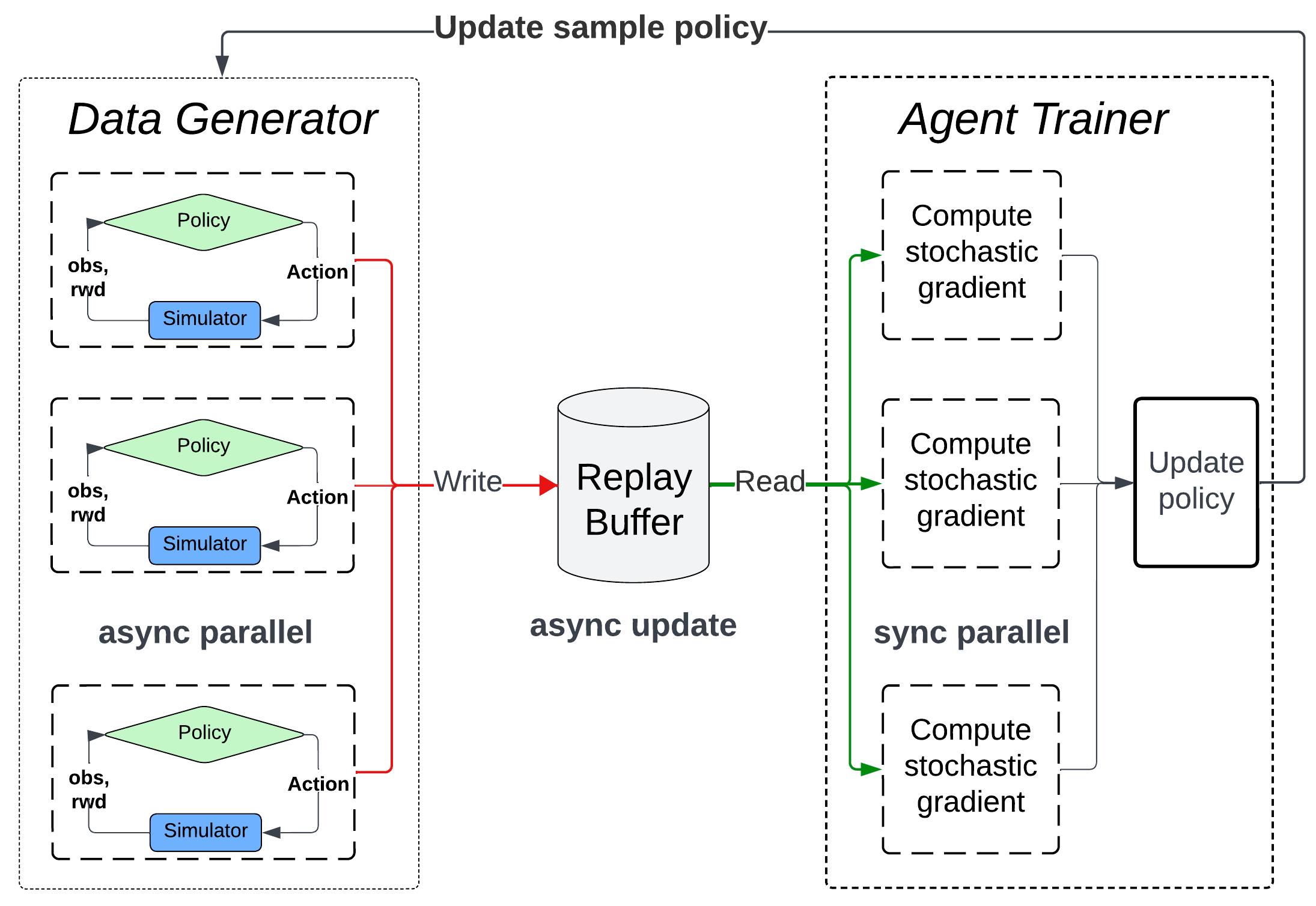}
    \caption{The high-level structure of RecoMind. The training consists of two components: a data generator and an RL trainer.}
    \label{fig:parallel}
    \vspace{-15pt} 
\end{figure}

To accelerate training, RecoMind adopts a scalable distributed architecture for synthetic data generation alongside distributed training, as illustrated in Figure \ref{fig:parallel}. This design decouples the data generation process from the policy training process, leveraging asynchronous parallelism to enhance efficiency and scalability.

In this distributed architecture, the data generator and trainer are interconnected through a replay buffer, facilitating continuous data flow and policy updates. The workflow is structured as follows:
\begin{itemize}[leftmargin=15pt]
\item \textbf{Data Generation}: Each generator instance loads the trained RL policy snapshot and interacts with the simulator to generate complete episode trajectories. Multiple generators run in parallel, collecting samples asynchronously and adding them to the replay buffer, ensuring a steady stream of training data.
\item \textbf{Policy Training}: The trainer continuously loads transition batches from the replay buffer to update the policy. A head node controls the training process by assigning mini-batches to training workers, who compute gradients in synchronous parallel. The head node aggregates these gradients and updates the policy. Policy snapshots are regularly published to the data generator to ensure it uses up-to-date policies for sampling.
\end{itemize}
The two components run continuously and asynchronously, in the sense that they do not wait for each other to proceed. Therefore, there could be some asynchronous delay: the data generator may not use the most recent snapshot for sampling and the trainer might update the policy with outdated transition samples. This trade-off is balanced by the increased data throughput and convergence speed.

Many existing RL libraries, such as RLlib \cite{liang2018rllib}, support such an infrastructure design, making it easier to implement at scale.

%% file: experiment.tex
\vspace{-5pt} 
\section{Experiments}\label{sec:experiment}
In this section, we present both offline and online experimental results to assess the effectiveness of the RecoMind RL framework in a video streaming recommendation task. We compare our approach to a widely used supervised learning based approach as this baseline reflects real-world industry practices and offers a meaningful benchmark to evaluate our framework's improvements.
\vspace{-10pt} 
\subsection{Simulator Setup}
As previously mentioned, the simulator in RecoMind uses a one-step greedy model to predict immediate feedback, as shown in Figure \ref{fig:simulator}. It uses three types of features: user embedding, item embedding, and user sequence. The user sequence consists of recent embeddings of recommended items and the corresponding user actions.

The user sequence is processed by transformer layers to capture correlations between the most recent recommended item embeddings and the corresponding user actions. It is then concatenated with user and item embeddings to form the input to a multilayer perceptron. The model outputs multiple heads, each predicting probabilities of different user actions such as save, watch, long watch (a watch of at least 10 seconds), hide, and exit. This model was trained on 700 million recent video feed sessions from over 80 million users and 100 million videos, where labels are formulated as a binary vector of the same dimension as the feedback set. 

As for the simulator, the initial user state is sampled from the dataset used for training the greedy model. The action set comprises items returned by the retrieval system. Once the agent selects an action, the simulator uses the one-step greedy model to predict the feedback for all possible user actions. The set $\cF_t$ is constructed with binary values, where an action is assigned a value of 1 if its prediction exceeds a predefined threshold (e.g. 0.5), and a value of 0 otherwise. This feedback set $\cF_t$ is then used to update the user state. Finally, the reward for the agent is derived from the prediction score associated with 'save' action, which serves as an indicator of user approval and interest. The detailed simulation and training process is outlined in Algorithm \ref{alg:sim_and_train}.

\vspace{-5pt} 
\subsection{RL Training Setup}
We use Deep Double $Q$-Learning \cite{ddql} to train the agent. The $Q$-network architecture is similar to the greedy one-step model's architecture but with an additional $|\cF|$-to-1 layer in the Multi-Layer Perceptron (MLP) to output a single $Q$-value (see Figure \ref{fig:DQN}). To effectively start the training, we bootstrap the $Q$-network with the greedy one-step model's pre-trained weights. Furthermore, we initialize the $|\cF|$-to-1 layer with a weight of 1 for 'save' action head and 0 otherwise. This setup allows the $Q$-network to pick up immediate rewards at the beginning of training.

In the exploration procedure, explained in detail in Section \ref{sec:framework}, the agent employs a custom method to selectively explore the highest-performing actions. Instead of relying on a static constant $K$, our method determines the top actions based on their performance percentage. To minimize computational overhead, this selection process occurs at the beginning of each episode only when the environment resets. The chosen action set remains fixed for all transitions within that episode. 

Furthermore, we integrate prioritized experience replay \cite{schaul2015prioritized}, which selects transitions for the mini-batch according to their temporal difference errors. This approach assigns higher sampling probabilities to transitions with larger errors, thereby emphasizing the most informative experiences during the training process. This prioritization allows the network to focus on significant experiences, thus improving sample efficiency. Additionally, we employ the n-step return technique to incorporate rewards from multiple future steps. This method improves the accuracy of the estimation of the value function, leading to better long-term reward predictions and better quality action selection.

Table \ref{tab:hyperparameter} lists the hyperparameters, including the Replay buffer $\alpha$, which controls the level of prioritization from 0 (no prioritization) to 1 (full prioritization), and the Replay buffer $\beta$, which adjusts the degree of bias correction from 0 (no correction) to 1 (full correction).

RecoMind was trained for two days using 8 NVIDIA A100 GPUs \cite{A100}. To ensure efficiency, we employed 1,000 data generator workers for parallel sampling and 8 RL training workers (one per GPU) for gradient computation.

\renewcommand{\arraystretch}{1.2}
\begin{table}[h!]  
\vspace{-5pt} 
\centering  
\begin{tabular}{l|c}  
\hline
\hline
\vspace{0.02in}
\textbf{Hyperparameter}& \textbf{Value}\\  
\hline  
\vspace{0.01in}
Prioritized exploration action percentage& 25\%\\
Softmax temperature&0.1\\
$\varepsilon$ &0.2\\
\hline 
Replay buffer size& 1000000\\  
Replay buffer $\alpha$&0.9\\
Replay buffer $\beta$ &0.1\\
\hline
Mini-batch size& 128\\  
Optimizer & Adam \cite{kingma2014adam} \\
$\gamma$ & 0.75\\
\hline
\end{tabular}  
\vspace{0.1in}
\caption{RecoMind Training Hyperparameters}  
\label{tab:hyperparameter}  
\vspace{-25pt} 
\end{table} 

\subsection{Offline Performance}\label{sec:exp_explore}

We evaluate the RL policy against a traditional greedy one-step model, which serves as the baseline. This baseline model predicts immediate feedback and selects items based on the highest immediate reward using the 'save' prediction head. In contrast, the RL policy selects items based on the highest $Q$ value. The 'save' head score is used in the baseline because it aligns with the RL model's reward function, ensuring a fair comparison between both policies.

Both policies are tested in a simulated environment over 100,000 episodes. The following metrics are chosen to measure different aspects of user engagement, capturing both short-term interactions and long-term satisfaction:
\begin{itemize}[leftmargin=15pt]
    \item \textbf{watch}: the total number of videos watched by the user, regardless of the duration. This metric indicates the session length.
    \item \textbf{long watch}: the number of videos watched for at least 10 seconds, suggesting a deeper level of interest.
    \item \textbf{save}: the number of videos the user saved, which signifies interest in the content, and long-term commitment to the platform.
    \item \textbf{hide}: the number of videos the user chose to hide, reflecting user disinterest or dissatisfaction with the content.
\end{itemize} 

The results are presented in Table \ref{tab:feedback_comparison} and demonstrate the superior performance of the RL policy. Compared with the one-step greedy policy, the RL policy leads to more engaging sessions ($+2.9\%$ in watch), deeper interest in content ($+9.1\%$ in long watch), increased approval and future interest ($+4.8\%$ in save), and reduced disinterest ($-2.5\%$ in hide). This highlights the effectiveness of optimizing for long-term rewards in improving various aspects of user engagement, as demonstrated through simulated user sessions.

\renewcommand{\arraystretch}{1.2}
\begin{table}[h!]  
\centering  
\begin{tabular}{l|c}  
\hline
\hline
\vspace{0.02in}
\textbf{User Feedback Type} & \textbf{Simulated Event Counts} \\  
\hline  
\vspace{0.01in}
watch & +2.9\%\\
long watch & +9.1\% \\  
save & +4.8\% \\  
hide & -2.5\% \\    
\hline
\end{tabular}  
\vspace{0.1in}
\caption{Offline performance comparison of the RecoMind RL policy versus the one-step greedy policy. The table shows the percentage change in various user feedback types, indicating increased user engagement with the RL policy, and shows the policy's effectiveness in optimizing in-session rewards.}  
\label{tab:feedback_comparison}  
\vspace{-20pt} 
\end{table} 
\subsection{Ablation Study}
We conduct a series of ablation studies to validate the effectiveness of each technique integrated into RecoMind. For these experiments, we employ a training setup mirroring the offline evaluation process except for two differences: data generation and evaluation were performed on a small subset of 100 users, and a candidate set of 100,000 items. This smaller scale is chosen due to the high computational cost and the  duration to train on a larger set of users.

Despite the smaller user set, we ensured a representative and random sampling of users, capturing diverse behaviors to maintain validity. This approach allows for rapid, and controlled experimentation, ensuring that each component is validated before scaling.

\begin{figure}
    \centering
    \includegraphics[scale=0.4]{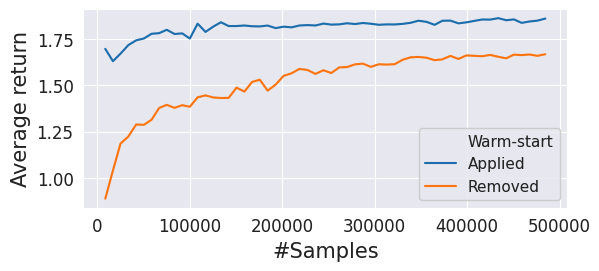}
    \vspace{-22pt} 
    \caption{Impact of $Q$-network warm-start. Applied represents training with a warm-started $Q$-network which can directly output immediate rewards at the beginning of training; Removed represents training with a randomly initialized $Q$-network with the same structure.}
    \label{fig:warm_start}
    \vspace{-10pt} 
\end{figure}

\textbf{Impact of Warm-Starting $Q$-network.} To showcase the benefit of this technique, we compare the warm-started $Q$-network with a randomly initialized $Q$-network while maintaining the same network structure. Figure \ref{fig:warm_start} illustrates the results. One can observe that after 500,000 iterations, the randomly initialized $Q$-network does not reach the performance level of the initial policy that leverages warm-starting. Furthermore, there is no guarantee that the randomly initialized policy will achieve the performance of the warm-started policy. This indicates significant time savings provided by the warm-start technique. 

\begin{figure}
    \centering
    \includegraphics[scale=0.4]{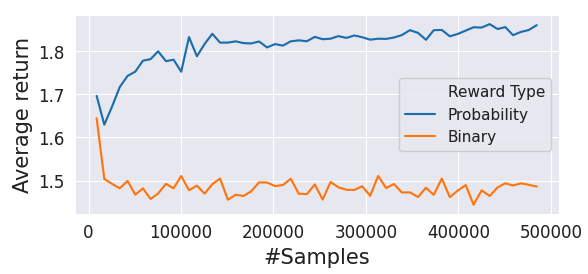}
    \vspace{-10pt} 
    \caption{Impact of reward function. Probability represents using feedback prediction scores as rewards and Binary represents using binary event indicators as rewards.}
    \label{fig:binary}
    \vspace{-10pt} 
\end{figure}

\textbf{Impact of Reward Function.} 
To investigate the impact of reward functions, we compare two reward definitions: feedback probability (a real number between 0 and 1) and an event indicator (0 or 1). The learning curves, presented in Figure \ref{fig:binary}, show that the feedback probability reward curve steadily improves, while the binary reward curve shows no gain after 500,000 samples. The binary reward proved to be sufficiently noisy, making it difficult for the RL agent to learn effectively.

\begin{figure}
    \centering
    \includegraphics[scale=0.35]{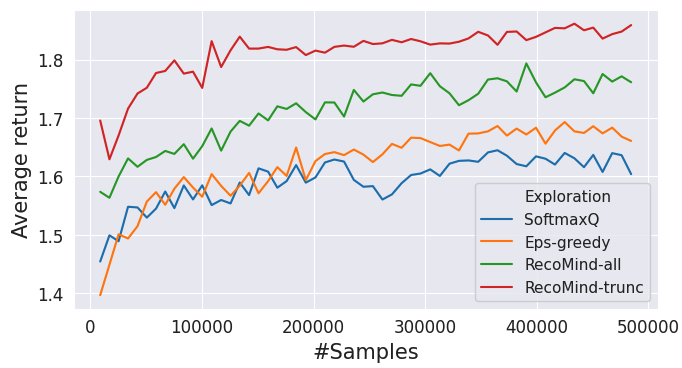}
    \vspace{-20pt} 
    \caption{Impact of exploration strategy. SoftmaxQ represents using softmax on $Q$-value to induce an action selection distribution; Eps-greedy represents $\varepsilon$-greedy; RecoMind-all represents RecoMind exploration without action truncation; RecoMind-trunc represents RecoMind exploration with $25\%$ prioritized action space truncation.}
    \label{fig:explore}
    \vspace{-15pt} 
\end{figure}

\textbf{Impact of Exploration Strategy}. 
To demonstrate the advantage of our custom exploration method, we compared four approaches: $\varepsilon$-greedy (Eps-greedy), softmax-$Q$ (SoftmaxQ), RecoMind custom exploration without action truncation (RecoMind-all), and RecoMind custom exploration with $25\%$ prioritized action space truncation (RecoMind-trunc), which was used in both offline and online experiments. The results, shown in Figure \ref{fig:explore}, reveal that RecoMind-trunc outperforms all others, with RecoMind-all ranked second. Eps-greedy and SoftmaxQ show rapid initial improvements but eventually hit a plateau. This highlights the effectiveness of RecoMind's exploration strategy in web-scale action spaces.

\vspace{-5pt} 
\subsection{Online Experiment}
To validate the effectiveness of RecoMind, we conducted an online A/B test on a video streaming platform, comparing the RL policy against a greedy one-step policy which serves as the baseline on 40 million users on our platform. Similar to the offline experiment, the RL policy recommended items with the highest $Q$-value prediction, while the one-step greedy policy recommended items with the highest immediate reward using the 'save' prediction head. Our RL policies were exclusively used at the ranking stage to select top items for the user from the candidate set provided by the retrieval stage. Although our solution has the capability to evaluate in-session value of every item in the catalog, for online inference, we focus only on a subset provided by the retrieval system. This approach, commonly adopted by recommendation systems, addresses online inference latency constraints effectively.

The results presented in Table \ref{tab:online_res} are statistically significant with $p$-values less than 0.01. In particular, there was an $8.85\%$ increase in the volume of videos watched and a $15.81\%$ in long watches (watch with at least 10 seconds), indicating both a higher volume of interaction and prolonged user engagement with the content.  Furthermore, the volume of saved videos increased by $3.69\%$, suggesting improved recommendation quality. The growth in all session depth metrics, including sustained increases in the "save" metric (a proxy for long-term commitment), further confirms the effectiveness of RL in achieving long-term rewards within the same session. Overall, the uplift across all measured metrics in the A/B testing underscores RecoMind's superior performance in enhancing user engagement and satisfaction compared to the existing one-step greedy baseline policy. These findings also confirm the results observed in offline evaluations.

\renewcommand{\arraystretch}{1.2}
\begin{table}[h!]  
\vspace{-5pt} 
\centering  
\begin{tabular}{l|crc}  
\hline
\hline
\vspace{0.02in}
\textbf{User Feedback Type} & \multicolumn{3}{c}{\textbf{Event Count \vs Baseline}} \\  
\hline  
\vspace{0.01in}
watch & &$+8.85\%$ \\
long watch & & $+15.81\%$ \\
save &  & $+3.69\%$\\
session depth $\geq 1$  & & $+1.74\%$\\
session depth $\geq 2$  & & $+3.11\%$\\
session depth $\geq 3$ & & $+3.52\%$\\
session depth $\geq 10$ & & $+4.71\%$\\
\hline
\vspace{0.01in}
\end{tabular}

\caption{RecoMind's Online Performance. The table shows the percentage increases in various types of user feedback when using the RecoMind RL policy compared to a baseline policy. These metrics indicate significant gains in user engagement and satisfaction, demonstrating the effectiveness of the RL policy in enhancing long-term user interactions.}  
\label{tab:online_res}  
\vspace{-30pt} 
\end{table} 

\vspace{-5pt} 
\subsection{Analysis of Diversity and Fairness}
 In addition to improving user engagement, we observed no change in diversity and fairness metrics, indicating that RecoMind does not introduce unwanted effects. In future work, we will further explore the relationship between exploration strategies and content diversity to identify methods that actively improves diversity while maintaining optimal user engagement.

%% file: conclusion.tex
\section{Conclusion}
In this paper, we introduce RecoMind, a simulator-based RL framework to optimize in-session engagement for web-scale recommendation systems. Our approach successfully combines the robustness of traditional one-step greedy models with a novel exploration strategy and the long-term optimization strengths of RL to effectively handle web-scale action spaces. Future research will focus on extending RecoMind to optimize across different sessions and longer horizons, as well as supporting multi-objective use cases.